# Pavement Image Datasets: A New Benchmark Dataset to Classify and Densify Pavement Distresses

**Hamed Majidifard, Corresponding Author**
Ph.D. Candidate
Civil and Environmental Engineering Department
University of Missouri-Columbia
Email: hmhtb@mail.missouri.edu

**Peng Jin**
Graduate Student
Civil and Environmental Engineering Department
University of Missouri-Columbia
Email: pj7r9@mail.missouri.edu

**Yaw Adu-Gyamfi**
Professor
Civil and Environmental Engineering Department
University of Missouri-Columbia
Email: adugyamfiy@missouri.edu

**William G. Buttlar**
Professor and Glen Barton Chair in Flexible Pavements
Civil and Environmental Engineering Department
University of Missouri-Columbia
Email: buttlarw@missouri.edu

**Abstract**

Automated pavement distresses detection using road images remains a challenging topic in the computer vision research community. Recent developments in deep learning has led to considerable research activity directed towards improving the efficacy of automated pavement distress identification and rating. Deep learning models require a large ground truth data set, which is often not readily available in the case of pavements. In this study, a labeled dataset approach is introduced as a first step towards a more robust, easy-to-deploy pavement condition assessment system. The technique is termed herein as the Pavement Image Dataset (PID) method. The dataset consists of images captured from two camera views of an identical pavement segment, i.e., a wide-view and a top-down view. The wide-view images were used to classify the distresses and to train the deep learning frameworks, while the top-down view images allowed calculation of distress density, which will be used in future studies aimed at automated pavement rating. For the wide view group dataset, 7,237 images were manually annotated and distresses classified into nine categories. Images were extracted using the Google Application Programming Interface (API), selecting street-view images using a python-based code developed for this project. The new dataset was evaluated using two mainstream deep learning frameworks: You Only Look Once (YOLO v2) and Faster Region Convolution Neural Network (Faster R-CNN). Accuracy scores using the F1 index were found to be 0.84 for YOLOv2 and 0.65 for the Faster R-CNN model runs; both quite acceptable considering the convenience of utilizing Google maps images.

# 1. INTRODUCTION

Recently, road infrastructure in the United States received a 'D' grade [1] according to the American Society of Civil Engineers (ASCE) Infrastructure Report Card, 2017. This is mainly driven by increasing budget constraints, which have created a culture of delayed maintenance and underinvestment in the renewal of transportation infrastructure systems. Strategic rehabilitation and maintenance of road surfaces require accurate data regarding overall road condition, and individual distress type, extent and severity [2]. Over the past two decades, improvement in sensor and camera technology has led to significant progress in automating pavement distress monitoring [3-5]. Typically, however, the sophisticated hardware and software involved results in high initial and operating expenses. For instance, a vehicle equipped with modern sensor and computing systems was purchased by the Ohio Department of Transportation for US$1,179,000, with an annual operating cost US$70,000 [6]. Furthermore, the final pavement condition assessment provided by these systems can be highly operator dependent [7].

Recent progress in image processing techniques and machine learning methods have motivated researchers to utilize these approaches to develop predictive models [8-10] towards well-timed repair and maintenance activities [2; 11]. Recent advances in deep learning has led to significant improvement of machine learning models in areas such as smart cities, self-driving cars, nanomaterials [12], transportation [13-14], healthcare [15], agriculture [16], retailing, and finance [17]. Similar strategies could be implemented in pavement distress monitoring. However, deep learning models rely on a large database of ground truth data, which is usually not available. Recent advances in vision-based, automated pavement crack detection techniques include: intensity-thresholding [18-21], match filtering [22], edge detection [23], seed-based approach [24], wavelet transforms [25-28], texture-analysis, and machine learning [29- 32]. An automatic procedure for crack detection known as CrackTree was reported by Zou et al. [33].

Although manipulation of machine learning methods for automated pavement distress detection is no longer a new technique, application of deep learning methods is still an area of active research [34-38]. Deep convolution neural networks (DCNNs) are defined as deep architecture with many hidden layers that enable them to acquire numerous abstraction levels [39-42]. Zhang et al. developed a crack detection model using raw image patches via the CNN-based software CrackNet [37]. In 2019, Zhang et al. implemented the Recurrent Neural Network (RNN) technique to create CrackNet-R, which is more efficient than CrackNet in detecting small cracks and in removing noise [43].

In order to develop a robust distress detection model, critical pavement distresses must be detected. Several factors such as traffic, climate, structural layering, layer age and condition will affect the pavement deterioration rate. Once characterized, road administrators can use the information to develop strategies to repair the pavement based on the type, extent and severity of the distresses identified. Previous studies made progress towards this goal, but falling short in one area or another. For instance, CrackNet [34] concentrated on determining the presence of damage, but did not specifically identify individual types of distress. Zalama et al. [44] categorized the types of distress horizontally and vertically, while Akarsu et al. [45] classified distresses into three categories – horizontal, vertical, and alligator. Finally, other investigations led to detection of blurry road markings [46], while other focused on classifying cracks, including sealed cracks [47].

The robustness of machine learning models is heavily dependent on the quality of data used for training them. Understanding the significance of labeled datasets for developing a robust pavement condition tool, in this study we introduce the so-called 'Pavement Image Dataset,' or (PID). The initial study utilized 7,237 images extracted from 22 different pavement sections, including both interstate and US highways. Images were extracted using the Google Application Programming Interface (API) in street-view using a code developed in Python. Initially, each image was hand-annotated by drawing a bounding box around each identified pavement distress. The dataset was evaluated using two classical deep learning frameworks, namely You Look Only Once (YOLO v2) and Faster Region Convolution Neural Network (Faster R-CNN). The following summarize the primary contributions of this study:

1. Introduction of a new dataset that enables simultaneous classification and density quantification of pavement distresses using varied camera views (top-down and wide-view). Wide-view images were used for classification, while top-down images were used for quantification of crack density.
2. Annotation of 7,237 images (wide-view images) with nine different distress types that were deemed to be critical for assessing pavement condition. These include a number of cracking modes, including reflective, transverse, block, longitudinal, alligator, sealed transverse, sealed longitudinal, and lane longitudinal cracking, along with potholes.
3. Implementation of two classical deep learning frameworks: YOLO-v2 and Faster R-CNN, and training of the models using the aforementioned dataset.

In the following section, we review previous datasets and introduce our proposed dataset.

## 2. PREVIOUS DATASETS

Several benchmarked datasets (private and public) have been developed in previous studies in the training of machine learning models [48]. The camera views typically used can be grouped into two categories: wide-view and top-down view. The main difference between the dataset provided in this study and previous datasets is that the current study captures data from both camera-views, which was found to be useful for distress classification and density determination. In the following sections, we review previous datasets based on either wide-view or top-down view datasets.

### 2.1. Wide-view datasets

Wide-view datasets capture a large area of the pavement and are therefore useful for pavement distress classification. Street-view image databases normally involve a high number of images with 'non pavement' views showing sidewalks, cars, buildings, etc. In a recent study, deep learning was employed to find and remove such objects based on a database of 9,712 wide-view images obtained via a mobile mapping system [49]. In another study, Maeda et al. used an end-to-end deep learning framework for pavement distress classification based on wide-view road images captured with a smartphone mounted on a vehicle dashboard. Images were divided into eight output classes (five types of cracks, rutting-bump-pothole-separation, crosswalk blur, and white line blur) [36]. Zhang et al. employed a sampling approach to create one million triple-channel (RGB) $99 \times 99$-pixel image patches based on 500 ($3264 \times 2448$ pixels) pavement images gathered by smartphone. In this study, 640,000, 160,000, and 200,000 patches were used for training, cross-validation, and testing, respectively [37].

### 2.2. Top down-view datasets

Top-down images provide more accurate view of distresses compared to wide-view. However, these type of images generally required more sophisticated camera and mounting equipment as compared to wide view images. In addition, pavement distress classification based on top-down views can be challenging as they may not capture the entire view of the distress. The German Asphalt Pavement distress (GAPs) dataset introduced by Eisenbach et al. was evidently the first open source pavement distress image dataset appropriate for high-performance DCNNs training. The study involved 1,969 grayscale pavement images (1,418 for training, 500 for testing, and 51 for validation) with different distresses such as cracks (alligator, sealed/filled longitudinal/transverse), patches, open joints, potholes, and bleeding [50]. In another study Gopalakrishnan et al. used a dataset containing over 1,000 pavement images provided from the Long-Term Pavement Performance (LTPP) database of the Federal Highway Administration (FHWA), which contained a combination of PCC-surfaced and AC-surfaced pavement images [51]. Zhang et al. used rotation data augmentation and image resizing methods to generate a large block dataset from 800 images. The idea of the research focused on classifying cracked, sealed and non-cracked blocks [47].

New image capturing technologies have recently been implemented to characterize pavement condition. Zhang et al. used an effective DCNN for pixel-perfect crack detection on three-dimensional asphalt pavement surfaces. The dataset included 1,800, 3D asphalt surface images for training, and another 200 images for testing the system [43]. Tong et al. utilized Ground Penetrating Radar (GPR) pavement images for automated identification, measurement, and detection of concealed cracks. The dataset contains 6,832 GPR images with different damage types such as subgrade settlement, hidden cracks, roadbed cavities, and non-damaged areas. GPR is a powerful technique for evaluating pavement integrity in a non-destructive manner, and can characterize subsurface pavement defects, such as hidden cracks [52].

Table 1 represents a summary of the datasets utilized in previous studies. Most of the studies relied on 2D images, while the Zhang et al. study utilized 3D asphalt pavement surface images. Public datasets have clearly aided in the development of open-source deep learning methods in pavement evaluation. This has also facilitated comparisons between models, for instance, in terms of their detection accuracy.

Table 1. Overview of datasets used in previous studies

| Availability | Reference | Dataset | Number of images | Number of classes | Angle of camera |
|---|---|---|---|---|---|
| Private | Some 2016 [49] | Street view images | 9,712 | 1 | wide view |
| | Zhang et al. 2016 [37] | Smartphone images | 500 | 1 | wide view |
| | Tong et al. 2017 [52] | GPR images | 6,832 | 3 | top-down |
| | Zhang et al. 2017 [34] | 3D asphalt surface images | 2,000 | 1 | top-down |
| | Zhang et al. 2018 [47] | Local (private) | 800 | 2 | top-down |
| Public | Eisenbach et al. 2017 [50] | German asphalt pavement distress (GAPs) | 1,969 | 6 | top-down |
| | Maeda et al. 2018 [36] | Smartphone street view images | 9,053 | 6 | wide view |
| | Fan et al. 2018 [38] | CFD; AigleRN | 157 | 1 | top-down |

| Gopalakrishnan et al. 2017 [51] | FHWA/LTPP | 1,056 | 1 | top-down |

All this notwithstanding, none of the aforementioned studies utilized a comprehensive dataset containing all pavement distress types from sections with highly varied condition. Furthermore, the studies did not attempt classification and distress density characterization in a simultaneous fashion. In the current study, we introduce a dataset with both wide-view and top-down view images to classify and determine distress density, respectively. Furthermore, we introduce a newly developed python-based software program, which was used to rapidly extract images in bulk within the Google API.

## 2. NEW DATASET

The current dataset consists of 7,237 images obtained from 22 different pavement sections in the United States collected by utilizing our new python-based code. The Google API enables the extraction of pavement images automatically by specifying GPS coordinates along with camera and image parameters. For each considered section, start and endpoints were selected on the road, and interpolated 'snapping points' were determined in 15 meter intervals. Two different images were collected at each coordinate point. Images with a pitch angle of -70° and -90° were chosen for distress classification and density determination, respectively. The wide view image (at a -70° pitch) was found to be useful for distress classification. The top-down view image (-90° pitch) led to more accurate distress quantification. Image size was 640×640 pixels for all images in the dataset. Afterwards, the wide-view images were hand annotated to characterize nine different distresses. Of the total 7,237 wide-view images, 5,789 images were used for training and 1,448 images were used for testing. The most critical distresses which affect pavement condition selected after reviewing various studies [53-59]. Afterward, the dataset was manually annotated using Openlabling software, which is a python-based software designed for object annotation [60]. The annotated dataset and the top-down images are available in the GitHub repository [61]. Figure 1 represents provides examples of the nine different distress types that were targeted.

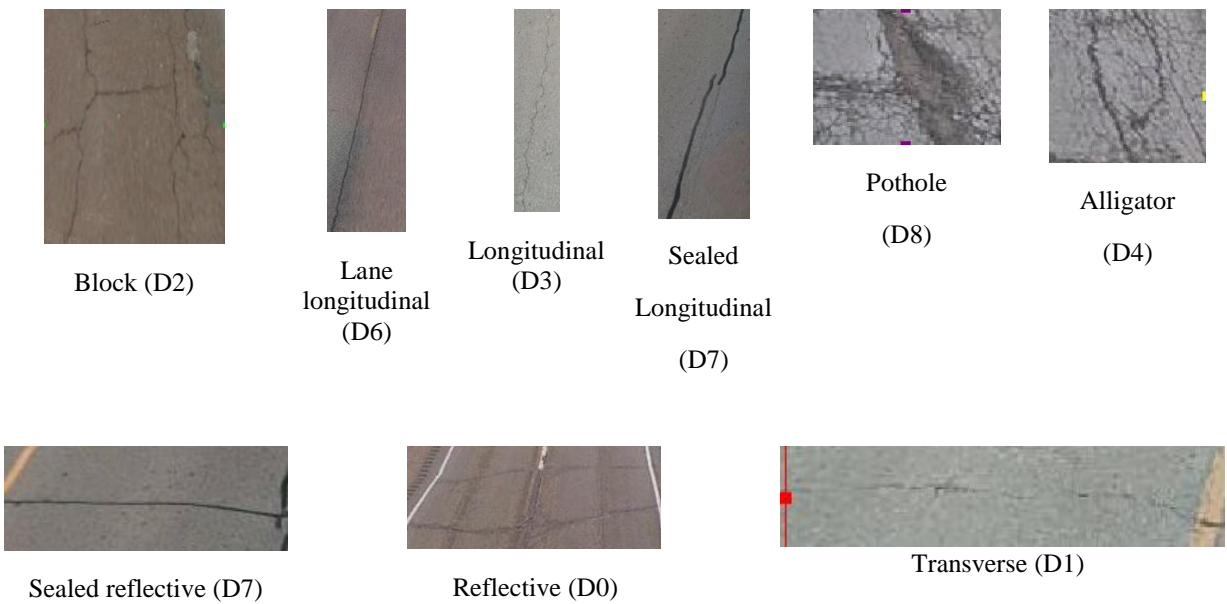

Block (D2)   Lane longitudinal (D6)   Longitudinal (D3)   Sealed Longitudinal (D7)   Pothole (D8)   Alligator (D4)

Sealed reflective (D7)   Reflective (D0)   Transverse (D1)

a)

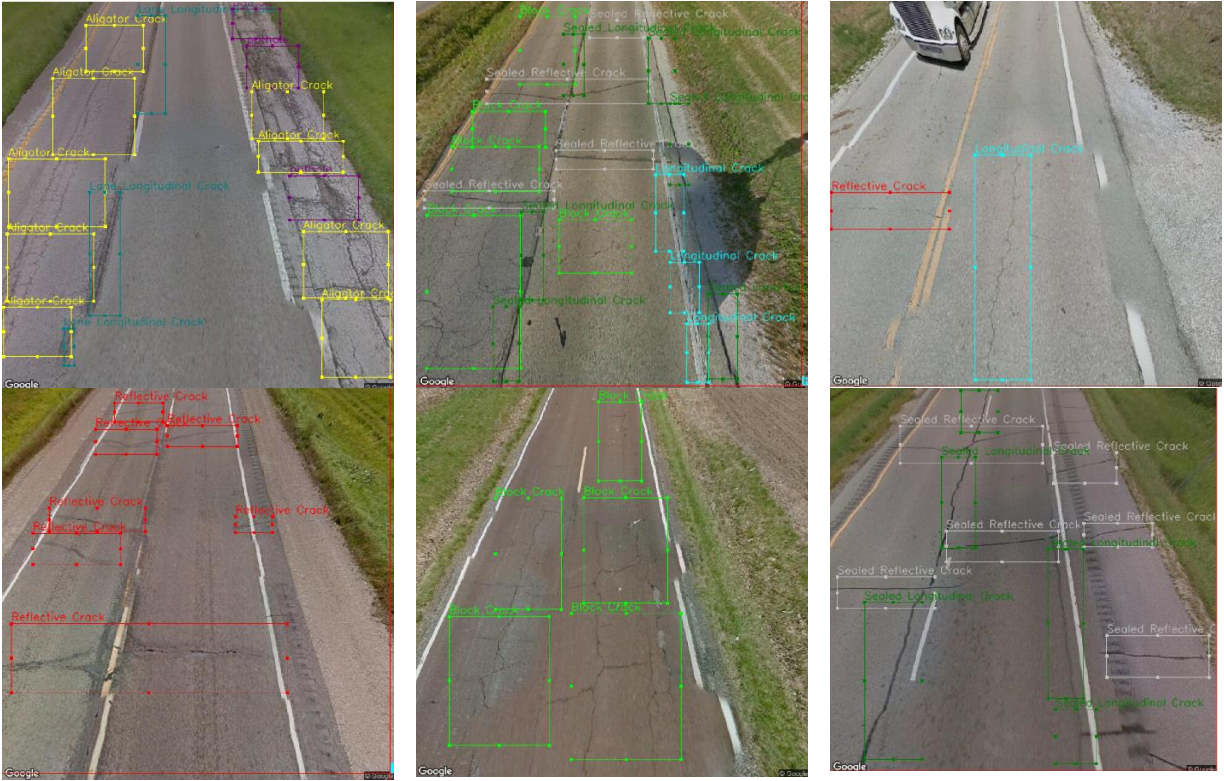

b)

Figure 1. a) Nine different distress classes considered in the PID dataset; b) sample of annotated images from wide-view images in the PID dataset.

The total number of boundary boxes and images for each distress type are shown in Figure 2. Reflective, lane longitudinal, sealed longitudinal, and block cracks are among the highest number of boundary boxes and images found in the selected pavements, mainly concentrated in the Midwest USA. Potholes were the scarcest distress found in our dataset, probably because our dataset focused on 'high-type' interstate and highways roads, where pothole repair is quickly done when needed.

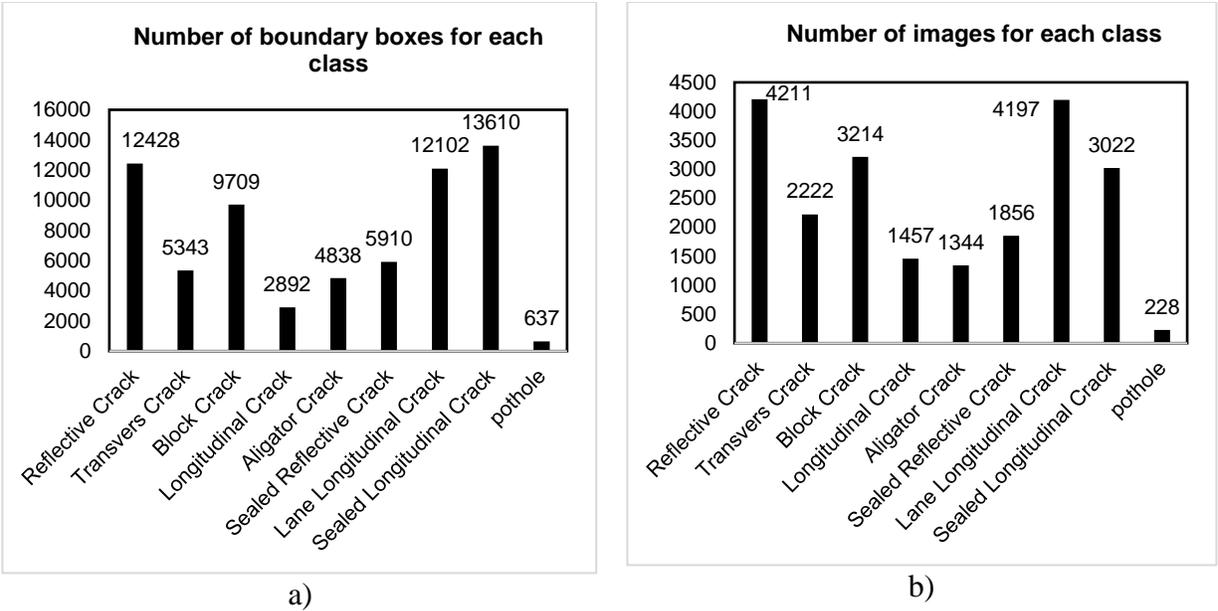

Figure 2. a) Number of boundary boxes for each class, b) Number of images for each class.

## 3. MODEL TRAINING AND TESTING

### 3.1. YOLO v2 Model

The first framework evaluated was the YOLO v2 deep convolutional neural network. YOLO is a relatively new object detection algorithm, which appears to have the highest accuracy and speed for developing deep learning-based models. YOLO reframes object detection methodology by looking at a particular image only one time to conduct object detections appropriately. Most recently, object detection algorithms use CNN classifiers to facilitate detections. In this manner, the algorithm can do simultaneous prediction of class probabilities. Table 2 shows the CNN architecture implemented for the prediction model developed herein. Standard layer types were used in the model including max pooling with a $2 \times 2$ kernel and convolution with a $3 \times 3$ kernel. The $1 \times 1$ kernel in the last convolutional layer contributes to reshape the data to $13 \times 13 \times 125$. This $13 \times 13$ structure is the size of the grid where the image becomes distributed. There are 35 channels of predictions for every grid cell. Through all these grid cells, five bounding boxes are predicted and labeled by seven data factors as follows: x and y values; height and width for the rectangle of the bounding box; road crack and non-crack probability distribution, and the confidence score.

Table 2. YOLO v2 model architecture.

| Layer | Kernel | Stride | Output Shape |
|---|---|---|---|
| Input | | | [416, 416, 3] |
| Convolution | 3×3 | 1 | [416, 416, 16] |
| Max Pooling | 2×2 | 2 | [208, 208, 16] |
| Convolution | 3×3 | 1 | [208, 208, 32] |
| Max Pooling | 2×2 | 2 | [104, 104, 32] |
| Convolution | 3×3 | 1 | [104, 104, 64] |
| Max Pooling | 2×2 | 2 | [52, 52, 64] |
| Convolution | 3×3 | 1 | [52, 52, 128] |
| Max Pooling | 2×2 | 2 | [26, 26, 128] |
| Convolution | 3×3 | 1 | [26, 26, 256] |
| Max Pooling | 2×2 | 2 | [13, 13, 256] |
| Convolution | 3×3 | 1 | [13, 13, 512] |
| Max Pooling | 2×2 | 1 | [13, 13, 512] |
| Convolution | 3×3 | 1 | [13, 13, 1024] |
| Convolution | 3×3 | 1 | [13, 13, 1024] |
| Convolution | 1×1 | 1 | [13, 13, 35] |

### 3.2. Faster R-CNN Model

The Faster R- CNN model involves a two-stage target detection method. The faster R-CNN model is a third generation model in the R- CNN series, which merges four primary steps in target detection. These steps include informative region selection, feature extraction classification, and location refinement into a deep network framework. It improves upon Fast R-CNN [62] by replacing the selective search method with a Region Proposal Network (RPN). First, Faster R-CNN splits an image into multiple, small segments. Next, the model passes each segment through a series of convolutional filters to derive the precious feature descriptors, which are subsequently passed through a classifier. The classifier outputs are the probability that an image area includes an object type. An NVIDIA GTX 1080Ti GPU was used to run this algorithm efficiently. The training time for the Faster R- CNN model was approximately four hours.

### 3.3. Transfer learning

Transfer learning was utilized to boost the training speed and performance of the YOLO and Faster R-CNN models. Using this method, a new task can benefit from formerly well-trained models. The Microsoft COCO dataset involves over 2 million well-labeled objects (like cars, shadows, etc) in 80 various groups with over 300,000 images. The pre-trained weights in the COCO dataset were used to initiate the detection task in the newly proposed models.

## 5. RESULTS

### 5.1. Model Accuracy

The performance of the proposed model was evaluated on 1,448 test images. The model was trained on 5,789 images for 40,000 iterations with the learning rate set to 0.01. The accuracy of the proposed model was evaluated by measuring the overlapping percentage between the ground truth and the prediction boxes. If a prediction box captured over 30% overlap with the ground truth box (Intersection over Union (IoU)), the prediction was considered a successful match, or a true positive (tp). Conversely, if the predicted bounding box had less than 30% IoU overlap with the ground truth box, it was categorized as a false positive (fp). Also, when there was an overlap of 30% between the prediction and the ground truth, but the predicted classification was incorrect, a false positive was assigned. False negatives (fn) were assigned to the instances where the model was not able to predict any distress.

Precision, Recall, and F1 score were the parameters used to evaluate model accuracy. Precision, shown in equation (1), is the ratio of true positives (tp) to all predicted positives (tp+fp). Correspondingly, Recall is the ratio of true positives to all the actual positives (tp+fn) as represented in equation (2). Overall accuracy is measured by the F1 score, which includes the recall values and a measure of statistical precision, as shown in equation (3).

$$Precision = \frac{tp}{tp+fp} \quad (1)$$

$$Recall = \frac{tp}{tp+fn} \quad (2)$$

$$F1 = \frac{2 \times Precision \times Recal}{Precision + Recal} \quad (3)$$

Figure 3 illustrates examples of detection and classification obtained using the YOLO algorithm. A red bounding box corresponds to the ground truth value, whereas a green bounding box represents the predicted value generated by the model. Figure 3(a) illustrates distresses that were accurately detected and classified with over 30% IoU for each crack class and thus, classified as true positive. Figure 3(b) illustrates a false positive (boundary box on the left), which has less than 30% IoU overlap with the ground truth. Distresses that were not detected by the model (False negative) are shown in Figure 3(b) and (c). Although some distresses were inadvertently left unlabeled during the tedious, manual annotation process, the model typically detected and classified them (Figure 3(d)). This suggest the high level of performance of the developed model within YOLO v2 algorithm.

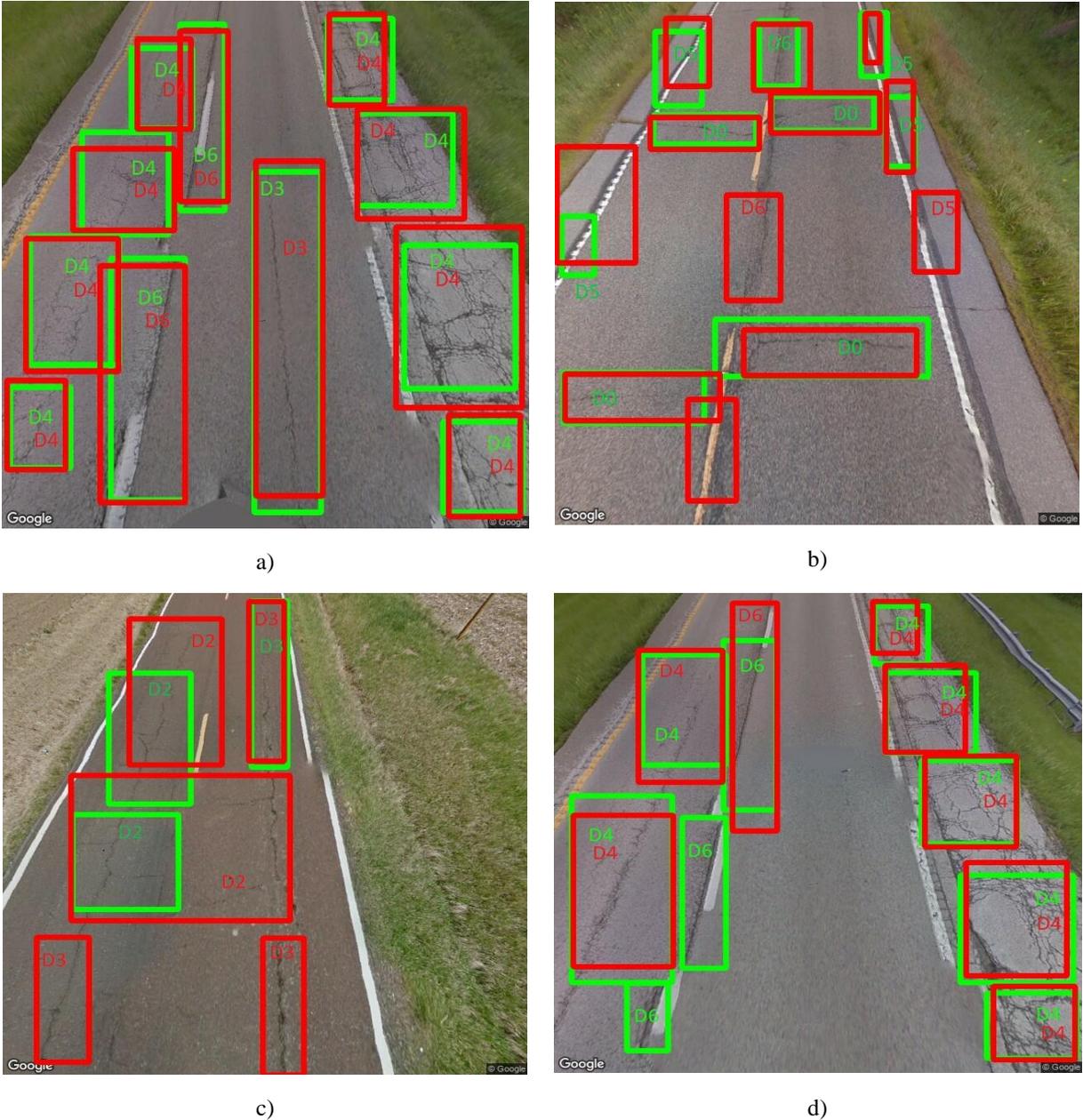

Figure 3. Classification of predicted crack from validation dataset: a) true positive, b) false positive and false negative, c) false negative, d) missed annotations identified by YOLO v2.

'Confusion matrices' from YOLO v2 and Faster R-CNN models are shown in Figure 4. Although the accuracy of both models was excellent, the YOLO v2 model achieved higher accuracy as indicated by the values in the confusion matrix as compared to the Faster R-CNN. In both models, confusion results occurred in a small, but significant number of cases. Relatively, confusions between classes occurred far more often in the Faster R-CNN model than the YOLO v2 model. Table 3 contains the confusions that were detected between classes in both models. Reflective and transverse cracks were the distresses most often confused in both models (Table 3). Also, alligator cracking and potholes were confused in several cases in both models. This can be explained by the similarity of these two distresses, as potholes often emerge as a later stage of alligator cracking.

| YOLO V2 | D0 | D1 | D2 | D3 | D4 | D5 | D6 | D7 | D8 |
|---|---|---|---|---|---|---|---|---|---|
| D0 | 0.99 | 0.01 | 0.00 | 0.00 | 0.00 | 0.01 | 0.00 | 0.00 | 0.00 |
| D1 | 0.02 | 0.97 | 0.01 | 0.00 | 0.00 | 0.00 | 0.00 | 0.00 | 0.00 |
| D2 | 0.00 | 0.00 | 0.99 | 0.00 | 0.00 | 0.00 | 0.00 | 0.00 | 0.00 |
| D3 | 0.00 | 0.00 | 0.01 | 0.98 | 0.00 | 0.00 | 0.01 | 0.00 | 0.00 |
| D4 | 0.00 | 0.00 | 0.00 | 0.00 | 0.99 | 0.00 | 0.00 | 0.00 | 0.00 |
| D5 | 0.00 | 0.00 | 0.00 | 0.00 | 0.00 | 1.00 | 0.00 | 0.00 | 0.00 |
| D6 | 0.00 | 0.00 | 0.00 | 0.00 | 0.00 | 0.00 | 0.99 | 0.00 | 0.00 |
| D7 | 0.00 | 0.00 | 0.00 | 0.00 | 0.00 | 0.00 | 0.00 | 1.00 | 0.00 |
| D8 | 0.00 | 0.00 | 0.00 | 0.00 | 0.00 | 0.00 | 0.00 | 0.00 | 1.00 |

a)

| Fast R-CNN | D0 | D1 | D2 | D3 | D4 | D5 | D6 | D7 | D8 |
|---|---|---|---|---|---|---|---|---|---|
| D0 | 0.96 | 0.02 | 0.01 | 0.00 | 0.00 | 0.01 | 0.00 | 0.00 | 0.00 |
| D1 | 0.05 | 0.91 | 0.04 | 0.00 | 0.00 | 0.00 | 0.00 | 0.00 | 0.00 |
| D2 | 0.00 | 0.01 | 0.97 | 0.02 | 0.00 | 0.00 | 0.00 | 0.00 | 0.00 |
| D3 | 0.00 | 0.00 | 0.07 | 0.92 | 0.00 | 0.00 | 0.01 | 0.00 | 0.00 |
| D4 | 0.00 | 0.00 | 0.00 | 0.00 | 0.97 | 0.00 | 0.01 | 0.00 | 0.01 |
| D5 | 0.01 | 0.00 | 0.00 | 0.00 | 0.00 | 0.99 | 0.00 | 0.00 | 0.00 |
| D6 | 0.00 | 0.00 | 0.00 | 0.01 | 0.00 | 0.00 | 0.99 | 0.00 | 0.00 |
| D7 | 0.00 | 0.00 | 0.00 | 0.00 | 0.00 | 0.00 | 0.01 | 0.99 | 0.00 |
| D8 | 0.00 | 0.00 | 0.00 | 0.00 | 0.07 | 0.00 | 0.00 | 0.00 | 0.93 |

b)

Figure 4. Confusion matrices obtained on the classification dataset using a) YOLO v2 and b) Fast R-CNN models

Table 3. Confusion between the classes in the models

| Model | Distress ID | Distresses | Number of Images |
|---|---|---|---|
| YOLO v2 | D0 and D1 | Reflective Crack and Transverse Crack | 19 |
| | D0 and D5 | Reflective Crack and Sealed Reflective Crack | 13 |
| | D1 and D2 | Transverse Crack and Block Crack | 6 |
| | D3 and D2 | Longitudinal Crack and | 8 |
| | D3 and D6 | Longitudinal Crack and Lane Longitudinal Crack | 4 |
| | D4 and D6 | Alligator Crack and Lane Longitudinal Crack | 7 |
| | D7 and D6 | Sealed Longitudinal Crack and Lane Longitudinal Crack | 7 |
| Faster R-CNN | D0 and D1 | Reflective Crack and Transverse Crack | 34 |
| | D0 and D2 | Reflective Crack and Block Crack | 11 |
| | D0 and D5 | Reflective Crack and Sealed Reflective Crack | 23 |
| | D1 and D2 | Transverse Crack and Block Crack | 27 |
| | D2 and D3 | Block Crack and Longitudinal Crack | 21 |
| | D3 and D6 | Longitudinal Crack and Lane Longitudinal Crack | 12 |
| | D4 and D6 | Alligator Crack and Lane Longitudinal Crack | 8 |

| | | |
|---|---|---|
| D4 and D8 | Alligator Crack and Pothole | 5 |
| D6 and D3 | Lane Longitudinal Crack and Longitudinal Crack | 12 |
| D7 and D6 | Sealed Longitudinal Crack and Lane Longitudinal Crack | 12 |

Table 4 shows detection and classification accuracies of the YOLO v2 and Faster R-CNN models for the nine classes in our dataset. Lower precision, recall, and F1 scores were found in the Faster R-CNN model in the cases of longitudinal, alligator, and longitudinal lane cracks. The F1 scores for the classes in the YOLO v2 model are higher than the scores for the Faster R-CNN model. The range of F1 scores in YOLO v2 models were between 0.95-0.98, while the Faster R-CNN model F1 scores ranged between 0.8-0.91. Although both are acceptable, the YOLO v2 model achieved better overall accuracy with an F1 score of 0.84 as compared to the Faster R-CNN model, which had an overall F1 score of 0.65. The precision and recall values for the YOLO v2 model were 0.93 and 0.77, respectively.

Among previous studies which used the same deep learning frameworks (YOLO v2), Maeda et al. model training resulted in precision and recall values of 0.77 and 0.71, respectively [36]. Mandal et al. achieved 0.77 and 0.73 for precision and recall values, respectively [39]. The high values of precision, recall and the F1 score of 0.84 in our proposed YOLO v2 model suggest the advantage of using labeled datasets in developing pavement distress detection models.

Table 4. Detection and classification results for nine distress types

| Crack class name | YOLO v2 | | | Faster R-CNN | | |
|---|---|---|---|---|---|---|
| | Precision | Recall | F1 | Precision | Recall | F1 |
| Reflective crack | 0.93 | 0.76 | 0.84 | 0.73 | 0.72 | 0.72 |
| Transverse crack | 0.9 | 0.83 | 0.86 | 0.75 | 0.74 | 0.75 |
| Block crack | 0.93 | 0.79 | 0.85 | 0.82 | 0.59 | 0.68 |
| Longitudinal crack | 0.91 | 0.83 | 0.87 | 0.66 | 0.43 | 0.52 |
| Alligator crack | 0.91 | 0.74 | 0.82 | 0.81 | 0.43 | 0.57 |
| Sealed transverse crack | 0.93 | 0.83 | 0.87 | 0.83 | 0.68 | 0.75 |
| Sealed longitudinal crack | 0.93 | 0.79 | 0.85 | 0.81 | 0.54 | 0.65 |
| Lane longitudinal crack | 0.94 | 0.57 | 0.71 | 0.75 | 0.3 | 0.42 |
| Pothole | 0.96 | 0.78 | 0.86 | 0.83 | 0.78 | 0.8 |
| **Average** | **0.93** | **0.77** | **0.84** | **0.78** | **0.58** | **0.65** |

**5.2. Model Performance when Using Top-Down Images**

In this section, models developed based solely on wide-view images were tested to allow comparison to results obtained using top-down view images. The motivation for training the models using wide-view images is due to the fact that wide view images are more readily obtained than top-down images. For instance, they can be obtained from smartphones in cars, whereas top-down images are harder to acquire as they emanate from more sophisticated equipment.

Figure 5 shows a comparison between the YOLO v2 and faster R-CNN frameworks in detecting pavement distresses from top-down images. Full sunshine images and images containing shadows (for instance, from trees) were selected in an attempt to challenge the robustness of each model. The black boxes in the Figure represents the ground truth, while the red, blue, and green boxes indicated the predicted detections. Both models were able to accurately detect distresses in both the full sunshine and shadow-containing images.

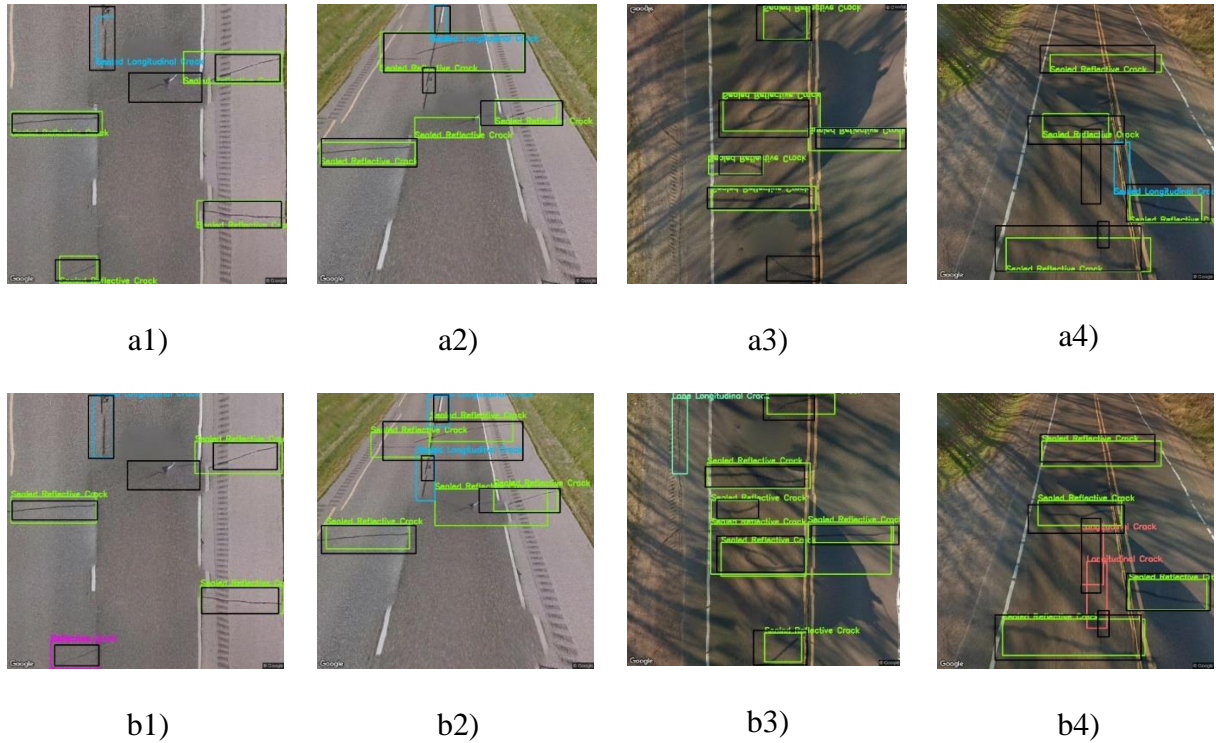

|  a1)  |  a2)  |  a3)  |  a4)  |

|  b1)  |  b2)  |  b3)  |  b4)  |

*The color of the bounding box indicates the type of distress and black is ground truth.

Figure 5. a) detection from YOLOv2 model b) detection from Faster R-CNN model, 1, 2, 3 and 4 represent plain top-down, plain wide-view, shadowed top-down and shadowed wide view image.

## 6. CONCLUSIONS

In this paper, we introduced a comprehensive dataset named Pavement Image Dataset (PID) for training machine learning models for the purpose of automated pavement distress characterization and monitoring. The dataset was created using Google API street-view via a python-based software, which was developed to extract pavement images at desired intervals along roadways. The dataset consists of two image groups: "wide-view images", where 7,237 images with bounding boxes featuring nine different pavement distresses were assembled; and "top-down images", consisting of 7,237 images at identical locations as the wide-view images. The wide-view images were used to classify distresses, while the top-down view images were used for calculating the density of distresses. The motivation for training the models using wide-view images is due to the fact that wide view images are more readily obtained than top-down images.

The primary focus of this article was to demonstrate how the wide-view images were used along with a deep learning approach to classify distresses. Two state-of-the-art, deep learning frameworks, YOLO v2 and Faster R-CNN, were implemented to automatically detect and classify nine types of pavement distress. The F1 scores, which are often used for model accuracy assessment, were obtained as 0.84 for YOLOv2 and 0.65 for the Faster R-CNN models, respectively. According to the F1 scores and confusion matrices for the nine distress classes, the YOLO v2 model results in more accurate distress characterizations than the Faster R-CNN model. The models developed based solely on wide-view images were tested on top-down view images to evaluate the ability of the model to detect distresses using different camera view images. Both models were able to accurately detect distresses in both the full sunshine, shadow-containing and car-containing images.

The proposed models offer some advantages over traditional pavement monitoring, and as compared to previous deep learning-based models. First, the models were trained using Google street-view images, which are free and available for virtually all roads in the US and abroad. Therefore, the model performance will be very accurate if Google street-view images are used for testing. Second, the models were developed based on a wide variety of common pavement distress types. The proposed dataset was annotated by pavement engineer experts for highway pavement sections, while the previous dataset focused on pavement distresses inside the city and were not annotated accurately. Finally, the developed models are robust and flexible, able to predict distress from different camera views towards convenient, cost-effective, and accurate pavement evaluation, monitoring, and management.

## 7. FUTURE WORK

In the current study, a pre-trained U-Net convolutional network was used, which was originally developed for biomedical image segmentation. Herein, it was used to quantify the density of cracks in roads [63]. The mentioned process was performed on top-down view images. Afterwards, the U-Net output image was reprocessed using a custom-developed MATLAB code to reduce image noise. The code is available in the GitHub repository [61]. Figure 7 shows the original, U-Net, and reprocessed image. Future research will be focused on developing improvements in the U-Net analysis. Also, the proposed model in YOLO and U-Net will be integrated with the python-based image extractor software developed in this study to grow the dataset directly from Google maps images. An automated estimate of Pavement Condition Index (PCI) could then be obtained [64], based on the number and amount of detected distress boxes and their intensities in each section. The accuracy of the new PCI parameter will then be validated by comparing results to 'foot-on-ground' manual inspection results, and a similar exercise could be done for other rating systems such as PASER [65].

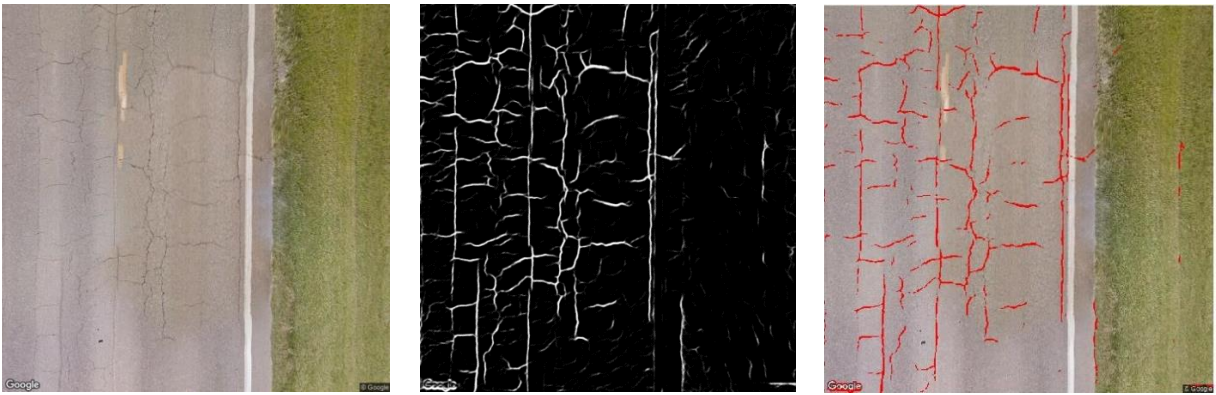

Figure 6. Examples of distress densification on top-down images. From left to right: original image, UNET output, overlapped modified UNET on original image

## 8. AUTHOR CONTRIBUTIONS

The authors confirm contribution to the paper as follows:

Study conception and design: Hamed Majidifard, Adu-Gyamfi, William Buttlar; data collection and annotation: Hamed Majidifard; Software setup and calibration: Adu-Gyamfi, Hamed Majidifard and Peng Jin: analysis and interpretation of results: Hamed Majidifard, Adu-Gyamfi; draft manuscript preparation: Hamed Majidifard, Adu-Gyamfi, William Buttlar. All authors reviewed the results and approved the final version of the manuscript.

## 9. REFERENCES

1. ASCE. *American Society of Civil Engineers (ASCE) 2017 Infrastructure Report Card: Roads*, American Society of Civil Engineers (ASCE), Reston, VA., USA, 2017.

2. Kargah-Ostadi, N., A. Nazef, J. Daleiden, and Y. Zhou. Evaluation Framework for Automated Pavement Distress Identification and Quantification Applications. *Transportation Research Record: Journal of the Transportation Research Board*, 2017. 2639(1): 46-54.

3. Paterson, W. D. Proposal of Universal Cracking Indicator for Pavements. *Transportation Research Record: Journal of the Transportation Research Board*, 1994. 1455: 69–75.

4. Lee, H. Accuracy, Precision, Repeatability, and Compatibility of a Pavedex PAS 1 Automated Distress Measuring Device. *Transportation Research Record: Journal of the Transportation Research Board,* 1991. 1311: 136–143.

5. KICT. *Final Report of The National Highway Pavement Management System 2016*. Gyeonggi-do, South Korea: Korea Institute of Civil Engineering and Building Technology, 2017.

6. Vavrik,W. R., L. D. Evans, J. A. Stefanski, and S. Sargand. *PCR Evaluation—Considering Transition from Manual to Semi-Automated Pavement Distress Collection and Analysis*. Columbus, OH: Ohio Department of Transportation, Office of Statewide Planning and Research, 2013.

7. Fu, P., J. T. Harvey, J. N. Lee, and P. Vacura. New Method for Classifying and Quantifying Cracking of Flexible Pavements in Automated Pavement Condition Survey. *Transportation Research Record: Journal of the Transportation Research Board*, 2011. 2225(1): 99-108.

8. Rashidi, M., M. Saghafi, and H. Takhtfiroozeh. Genetic Programming Model for Estimation of Settlement in Earth Dams. *International Journal of Geotechnical Engineering*, 2018. DOI:10.1080/19386362.2018.1543100.

9. Fathi, A., M. Mazari, and M. Saghafi. Multivariate Global Sensitivity Analysis of Rocking Responses of Shallow Foundations under Controlled Rocking. Eighth International Conference on Case Histories in Geotechnical Engineering, ASCE Geo-Congress, Philadelphia*,* PA., 2019. 307: 490-498, DOI: https://doi.org/10.1061/9780784482094.045.

10. Saghafi, M., S. M. Asgharzadeh, A. Fathi, and A. Hosseini. Image Processing Method to Estimate the Wearing Condition of Slurry Seal Mixtures. Transportation & Development Institute (T&DI), International Airfield and Highway Pavements Conference, Chicago, IL., 2019. https://doi.org/10.1061/9780784482452.042.

11. Jia, X., B. Huang, Q. Dong, D. Zhu, and J. Maxwell. Influence of Pavement Condition Data Variability on Network-Level Maintenance Decision. *Transportation Research Record: Journal of the Transportation Research Board*, 2016. *2589*(1): 20-31.

12. Hajilounezhad, T., Oraibi, Z. A., Surya, R., Bunyak, F., Maschmann, M. R., Calyam, P., & Palaniappan, K. Exploration of Carbon Nanotube Forest Synthesis-Structure Relationships Using Physics-Based Simulation and Machine Learning.2019.

13. Adu-Gyamfi, Y. O., Asare, S. K., Sharma, A., & Titus, T. Automated vehicle recognition with deep convolutional neural networks.*Transportation Research Record: Journal of the Transportation Research Board*, 2017. *2645*(1), 113-122.

14. Chakraborty, P., Adu-Gyamfi, Y. O., Poddar, S., Ahsani, V., Sharma, A., & Sarkar, S. Traffic congestion detection from camera images using deep convolution neural networks. *Transportation Research Record: Journal of the Transportation Research Board*. 2018. *2672*(45), 222-231.

15. Miotto, R., Wang, F., Wang, S., Jiang, X., & Dudley, J. T. Deep learning for healthcare: review, opportunities and challenges. *Briefings in bioinformatics*, 2017. *19*(6), 1236-1246.

16. Kamilaris, A., & Prenafeta-Boldú, F. X. Deep learning in agriculture: A survey. *Computers and electronics in agriculture*, 2018. *147*, 70-90.

17. Heaton, J. B., Polson, N. G., & Witte, J. H. Deep learning for finance: deep portfolios. *Applied Stochastic Models in Business and Industry*, 2017. *33*(1), 3-12.

18. Chambon, S., and J.-M. Moliard. Automatic Road Pavement Assessment with Image Processing: Review and Comparison. *International Journal of Geophysics,* 2011. 2011: 20.

19. Oliveira, H., and P. L. Correia. Automatic Road Crack Segmentation using Entropy and Image Dynamic Thresholding. 17th European Signal Processing Conference, Glasgow, 2009. pp. 622–626.

20. Tsai, Y. -C., V. Kaul, and R. M. Mersereau. Critical Assessment of Pavement Distress Segmentation Methods. *Journal of Transportation Engineering,* 2009. 136: 11–19.

21. Zhang, D., Q. Li, Y. Chen, M. Cao, L. He, and B. Zhang. An Efficient and Reliable Coarse-to-Fine Approach for Asphalt Pavement Crack Detection. *Image and Vision Computing*, 2017. 57: 130–146.

22. Zhang, A., Q. Li, K. C. Wang, and S. Qiu. Matched Filtering Algorithm for Pavement Cracking Detection. *Transportation Research Record: Journal of the Transportation Research Board*, 2013. 2367(1): 30-42.

23. Ayenu-Prah, A., and N. Attoh-Okine. Evaluating Pavement Cracks with Bidimensional Empirical Mode Decomposition. *EURASIP Journal on Advances in Signal Processing,* 2008. 2008: 861701, https://doi.org/10.1155/2008/861701.

24. Zhou, Y., F. Wang, N. Meghanathan, and Y. Huang. Seed-Based Approach for Automated Crack Detection from Pavement Images.*Transportation Research Record: Journal of the Transportation Research Board*, 2016. *2589*(1): 162-171.

25. Zhou, J., P. Huang, and F. P. Chiang. Wavelet-Based Pavement Distress Classification. *Transportation Research Record: Journal of the Transportation Research Board*, 2005. 1940(1): 89-98.

26. Subirats, P., J. Dumoulin, V. Legeay, and D. Barba. Automation of Pavement Surface Crack Detection using the ContinuousWavelet Transform. Proceedings of the 2006 International Conference on Image Processing, GA., 2006. pp. 3037–3040.

27. Wang, K. C. P., Q. Li, and W. Gong. Wavelet-Based Pavement Distress Image Edge Detection with à Trous Algorithm. *Transportation Research Record: Journal of the Transportation Research Board*, 2007. 2024: 73–81.

28. Ying, L., and E. Salari. Beamlet Transform-Based Technique for Pavement Crack Detection and Classification. *Computer-Aided Civil and Infrastructure Engineering*, 2010. 25: 572–580, https://doi.org/10.1111/j.1467-8667.2010.00674.x.

29. Koch, C., K. Georgieva, V. Kasireddy, B. Akinci, and P. Fieguth. A Review on Computer Vision Based Defect Detection and Condition Assessment of Concrete and Asphalt Civil Infrastructure. *Advanced Engineering Informatics,* 2015. 29: 196–210.

30. Oliveira, H., and P. L. Correia. Automatic Road Crack Detection and Characterization. *IEEE Transactions on Intelligent Transportation Systems,* 2013. 14: 155–168, DOI:10.1109/TITS.2012.2208630.

31. Fujita, Y., K. Shimada, M. Ichihara, and Y. Hamamoto. A Method Based on Machine Learning Using Hand-Crafted Features for Crack Detection from Asphalt Pavement Surface Images. Thirteenth International Conference on Quality Control by Artificial Vision, Tokyo, 2017. 103380I, https://doi.org/10.1117/12.2264075.

32. Hizukuri, A., and T. Nagata. Development of a Classification Method for a Crack on a Pavement Surface Images Using Machine Learning. Thirteenth International Conference on Quality Control by Artificial Vision*,* Tokyo, 2017. 103380M, https://doi.org/10.1117/12.2266911.

33. Zou, Q., Y. Cao, Q. Li, Q. Mao, and S. Wang. CrackTree: Automatic Crack Detection from Pavement Images. *Pattern Recognition Letters,* 2012. 33(3): 227-238.

34. Zhang, A., K. C. P. Wang, B. Li, E. Yang, X. Dai, Y. Peng, Y. Fei, Y. Liu, J. Q. Li, and C. Chen, Automated Pixel-Level Pavement Crack Detection on 3D Asphalt Surfaces Using a Deep-Learning Network. *Computer-Aided Civil and Infrastructure Engineering*, 2017. 32(10): 805-819.

35. Cha, Y.-J., W. Choi, and O. Büyüköztürk. Deep Learning-Based Crack Damage Detection Using Convolutional Neural Networks. *Computer-Aided Civil and Infrastructure Engineering,* 2017. 32: 361–378.

36. Maeda, H., Y. Sekimoto, T. Seto, T. Kashiyama, and H. Omata. Road Damage Detection Using Deep Neural Networks with Images Captured Through a Smartphone. *arXiv*, 2018.

37. Zhang, L., F. Yang, Y. D. Zhang, and Y. J. Zhu. Road Crack Detection Using Deep Convolutional Neural Network. IEEE International Conference on Image Processing (ICIP), Phoenix, AZ., 2016. pp. 3708-3712, DOI: 10.1109/ICIP.2016.7533052.

38. Fan, Z., Y. Wu, and W. Li. Automatic Pavement Crack Detection Based on Structured Prediction with the Convolutional Neural Network. *arXiv*, 2018.

39. Mandal, V., L. Uong, and Y. Adu-Gyamfi. Automated Road Crack Detection Using Deep Convolutional Neural Networks. IEEE International Conference on Big Data, WA., 2018. pp. 5212-5215, DOI: 10.1109/BigData.2018.8622327.

40. Xie, D., L. Zhang, and L. Bai. Deep Learning in Visual Computing and Signal Processing, *Applied Computational Intelligence and Soft Computing,* 2017. 2017: 13.

41. Agrawal, A., and A. Choudhary. Perspective: Materials Informatics and Big Data: Realization of the 'Fourthparadigm' of Science in Materials Science. *APL Materials*, 2016, 053208 https://doi.org/10.1063/1.4946894.

42. Liu, W., Z. Wang, X. Liu, N. Zeng, Y. Liu, and F. E. Alsaadi. A Survey of Deep Neural Network Architectures and Their Applications. *Neurocomputing*, 2017. 234: 11–26.

43. Zhang, A., K. C. P. Wang, Y. Fei, Y. Liu, C. Chen, G. Yang, J. Q. Li, E. Yang, and S. Qiu. Automated Pixel-Level Pavement Crack Detection on 3D Asphalt Surfaces with a Recurrent Neural Network. *Computer-Aided Civil and Infrastructure Engineering*. 2019. 34(3): 213-229.

44. Zalama, E., J. Gómez-García-Bermejo, R. Medina, and J. Llamas. Road Crack Detection Using Visual Features Extracted by Gabor Filters. *Computer-Aided Civil and Infrastructure Engineering*. 2014. 29(5): 342–358.

45. Akarsu, B., M. Karak ̈ose, K. Parlak, A. K. I. N. Erhan, and A. Sarimaden. A Fast and Adaptive Road Defect Detection Approach Using Computer Vision with Real Time Implementation, *International Journal of Applied Mathematics, Electronics and Computers*. 2016. 4: 290–295.

46. Kawano, M., K. Mikami, S. Yokoyama, T. Yonezawa, and J. Nakazawa. Road Marking Blur Detection with Drive Recorder, Proceedings of the 2017 IEEE International Conference on Big Data (Big Data), Boston, MA.2017. DOI: 10.1109/BigData.2017.8258427.

47. Zhang, K., H. D. Cheng, and B. Zhang. Unified Approach to Pavement Crack and Sealed Crack Detection Using Preclassification Based on Transfer Learning. *Journal of Computing in Civil Engineering.* 2018. 32: https://doi.org/10.1061/(ASCE)CP.1943-5487.0000736.

48. Gopalakrishnan, K. Deep Learning in Data-Driven Pavement Image Analysis and Automated Distress Detection: A Review. *Data*. 2018. *3*(3): 28, https://doi.org/10.3390/data3030028.

49. Some, L. Automatic Image-Based Road Crack Detection Methods, Thesis at KTH Royal Institute of Technology: Stockholm, Sweden, 2016.

50. Eisenbach, M., R. Stricker, D. Seichter, K. Amende, K. Debes, M. Sesselmann, D. Ebersbach, U. Stoeckert, and H. M. Gross. How to Get Pavement Distress Detection Ready for Deep Learning? A Systematic Approach. International Joint Conference on Neural Networks (IJCNN), Anchorage, AK., 2017. pp. 2039–2047.

51. Gopalakrishnan, K., S. K. Khaitan, A. Choudhary, and A. Agrawal. Deep Convolutional Neural Networks with Transfer Learning for Computer Vision-Based Data-Driven Pavement Distress Detection. *Construction and Building Materials*. 2017. 157: 322–330.

52. Tong, Z., J. Gao, and H. Zhang. Recognition, Location, Measurement, and 3D Reconstruction of Concealed Cracks Using Convolutional Neural Networks. *Construction and Building Materials*. 2017. 146: 775–787.

53. McGhee, K. Development and implementation of pavement condition indices for the Virginia Department of Transportation. *Phase I Flexible Pavements*. 2002.

54. Buttlar, W., J. Meister, B. Jahangiri, and H. Majidifard. *Performance Characteristics of Modern Recycled Asphalt Mixes in Missouri, Including Ground Tire Rubber, Recycled Roofing Shingles, and Rejuvenators*. 2019. No. cmr 19-002., Missouri DOT.

55. Buttlar, W., Rath, P., Majidifard, H., Dave, E. V., & Wang, H. Relating DC (T) Fracture Energy to Field Cracking Observations and Recommended Specification Thresholds for Performance-Engineered Mix Design. *Asphalt Mixtures*, 2018. Vol. 51.

56. Jahangiri, B., Majidifard, H., Meister, J., Buttlar, W. Performance Evaluation of Asphalt Mixtures with Reclaimed Asphalt Pavement and Recycled Asphalt Shingles in Missouri. *Transportation Research Record: Journal of the Transportation Research Board*. 2019, DOI: 10.1177/0361198119825638.

57. Wang, Y., Ghanbari, A., Underwood, B.S., Kim, Y.R. Development of a Performance-volumetric Relationship for Asphalt Mixtures. *Transportation Research Record: Journal of the Transportation Research Board*, 2019. Washington, D.C.

58. Ghanbari, A., Underwood, B. S., & Kim, Y. R. Development of Rutting Index Parameter Based on Stress Sweep Rutting Test and Permanent Deformation Shift Model. International Journal of Pavement Engineering, In Press.

59. Keshavarzi, B., Ghanbari, A., & Kim, Y. R., Extrapolation of Dynamic Modulus Data for Asphalt Concrete at Low Temperatures, Road Materials and Pavement Design, In Press.

60. Open Labeling Tool (2018). Available: https://github.com/Cartucho/OpenLabeling.git, [Accessed: 05-Dec-2019].

61. PID - Pavement Image Dataset (2019). Available: https://github.com/hmhtb/PID-Pavement-Image-Dataset.git, [Accessed: 05-Dec-2019].

62. Girshick, R. Fast R-CNN. *arXiv,* 2015. [Online]. Available: https://arxiv.org/abs/1504.08083. [Accessed: 30-Jul-2019].

63. Ronneberger, O., P. Fischer, and T. Brox. U-net: Convolutional Networks for Biomedical Image Segmentation. *arXiv*, 2015. 1505.04597.

64. Shahin, M. Y. *Pavement Management for Airports, Roads, and Parking Lots*. New York: Springer, 2005. Vol. 501.

65. Walker, D., L. Entine, and S. Kummer. *Asphalt PASER Manual Report: Pavement Surface Evaluation and Rating (PASER) Manuals.* Transportation Information Center, University of Wisconsin-Madison, 2002.